\documentclass{article}

\PassOptionsToPackage{numbers}{natbib}

\usepackage[final]{nips_2017}

\usepackage[utf8]{inputenc} 
\usepackage[T1]{fontenc}    
\usepackage{hyperref}       
\usepackage{url}            
\usepackage{booktabs}       
\usepackage{amsfonts}       
\usepackage{nicefrac}       
\usepackage{microtype}      

\usepackage{amsmath}
\usepackage{graphicx}
\usepackage{subcaption}
\usepackage[ruled,linesnumbered]{algorithm2e}
\usepackage{array}
\usepackage{multirow}
\usepackage{color}

\title{Actor-Critic Sequence Training for Image Captioning}

\author{Li Zhang \\   
Queen Mary University of London \\
{\tt\small david.lizhang@qmul.ac.uk }\\
\And
Flood Sung \\   
Independent Researcher\\
\texttt{floodsung@gmail.com}
\And
Feng Liu \quad Tao Xiang  \quad Shaogang Gong \quad Yongxin Yang\\
Queen Mary University of London \\
{\tt\small \{feng.liu, t.xiang, s.gong, yongxin.yang\}@qmul.ac.uk}\\
\And
Timothy M. Hospedales\\
 \quad The University of Edinburgh\\
{\tt\small  t.hospedales@ed.ac.uk}
}

\begin{document}

\maketitle

\begin{abstract}
Generating natural language descriptions of images is an important capability for a robot or other visual-intelligence driven AI agent that may need to communicate with human users about what it is seeing. Such image captioning methods are typically trained by maximising the likelihood of  ground-truth annotated caption given the image. While simple and easy to implement, this approach does not directly maximise the language quality metrics we care about such as CIDEr. In this paper we investigate training image captioning methods based on actor-critic reinforcement learning in order to directly optimise non-differentiable quality metrics of interest. By formulating a per-token advantage and value computation strategy  in this novel reinforcement learning based captioning model,
we show that it is possible to achieve the state of the art performance on the widely used MSCOCO benchmark. 
\end{abstract}

\section{Introduction}

As the classic task of automatic object category recognition is beginning to approach a solved problem \cite{szegedy2016rethinking}, interest is growing in solving a more `end-to-end' task of generating richer descriptions of images in terms of natural language, suitable for communication to human users \cite{vinyals2015show,vinyals2016show, you2016image, liu2016semantic}. This task is extremely topical recently, benefiting from public benchmarks such as MSCOCO \cite{lin2014microsoft}. Despite extensive research in recent years, leading performance on the benchmarks has not increased dramatically. 
We hypothesise that this is mainly due to research focus being on the image understanding aspects of captioning, rather than the language generation aspects, and in this paper investigate reinforcement learning methods for training effective language generation in captioning.

Most existing captioning studies investigate variants of deep learning-based image encoders, that feed into deep sentence decoders. They have two main issues: (i) They are trained by maximising the likelihood of each ground-truth word given the previous ground-truth words and the image using back-propagation \cite{ranzato2015sequence}, termed `Teacher-Forcing' \cite{bengio2015scheduled}.  This creates a mismatch between training and testing, since at test-time the model uses the previously generated words from the model distribution to predict the next word. This exposure bias \citep{ranzato2015sequence}, results in error accumulation during generation at test time, since the model has never been exposed to its own predictions. (ii) While sequence models are usually trained using the cross entropy loss, the actual NLP quality metrics of interest -- with which we evaluate them at test time -- are non-differentiable metrics such as CIDEr~\citep{vedantam2015cider}. Ideally sequence models for image captioning should be trained to avoid exposure bias and directly optimise metrics for the task at hand.

In this paper we propose to improve image captioning by addressing the above identified two issues through training captioning models with reinforcement learning. In this way, we can optimise the gradient of the expected reward by sampling from the model during training, thus avoiding the train-test mismatch; and we can directly optimise the relevant test-time metrics such as CIDEr, by treating them as reward in a reinforcement learning context.

Specifically we propose an actor-critic model for image captioning. It consists of a policy network (actor) and value network (critic). The actor is trained to predict the caption as a sequential decision problem given the image,  where the sequence of actions correspond to tokens. The critic predicts the value of each state \textcolor{black}{(image and sequence of actions so far)}, which we define as the expected task-specific reward (language metric score) that the network will receive if it outputs the current token and continues to sample outputs according to its probability distribution. The value predicted by the critic can be used to train the actor (captioning policy network). Under the assumption that the critic produces the exact values, the actor is trained based on an unbiased estimate of the gradient of the caption score in terms of relevant language quality metrics. Compared to most reinforcement learning applications \cite{mnih2013playing}, image captioning has a much higher dimension action (e.g., 10,000+ token/word actions) space but shorter episodes. \textcolor{black}{The proposed actor-critic approach exploits the shorter episodes and ameliorates the high dimensional action space.}  Our model achieves state-of-the-art performance: It is ranked third on the MS-COCO testing server leaderbord when submitted, which is the highest rank achieved by reinforcement learning based method without model ensemble.

\section{Related Work}

\noindent \textbf{Image captioning}\quad There is now extensive work on image captioning \cite{vinyals2015show, fang2015captions, xushow, devlin2015language, mao2014deep, wu2016value, vinyals2016show, liu2016semantic}. The typical pipeline is based on a convolutional neural network (CNN) image encoder and a recurrent neural network (RNN) based sentence decoder \cite{vinyals2015show,vinyals2016show}. Topical issues addressed in the literature include dynamic attention \cite{you2016image,xushow}, improving visual feature representations \cite{liu2016semantic}. As mentioned earlier they are typically trained by maximising training caption likelihood through teacher forcing. 

\noindent \textbf{Image captioning with reinforcement learning}\quad Recently a few studies proposed to use reinforcement learning to address the discrepancy between the standard training objective for image captioning (likelihood/teacher forcing) and the evaluation metrics of interest (CIDEr) \citep{ranzato2015sequence,liu2016optimization, rennie2016self}. 
\citep{rennie2016self} \textcolor{black}{uses the basic REINFORCE algorithm~\citep{williams1992reinforce} with a reward obtained by the current model under the inference algorithm as the baseline. As a results, for each sampled caption, it has only one sentence level advantage which means that every token makes the same contribution towards the whole sentence -- a clearly invalid assumption.}
\citep{ranzato2015sequence, liu2016optimization} \textcolor{black}{add an additional FC layer on top of the RNN output to predict state value function. However, both  treat \emph{state} as the RNN output while we treat the \emph{state} as the RNN input (given image and the taken actions), so that we can build an independent value network rather than a shared RNN cell between actor and critic.} 

\noindent \textbf{Actor-Critic} \quad  An actor-critic~\citep{Barto1983Neuronlike} method trains the actor by policy gradient with advantages baselined by the critic. Many state-of-the-art reinforcement learning algorithms~\citep{mnih2016asynchronous} are based on actor-critic. For instance, AlphaGo~\citep{silver2016mastering} utilised the actor-critic method to do self-learning in the game of Go and achieved great success by beating human world champions. It uses Monte-Carlo rollout and the reward is only set at the end of the game with very long episode.

\noindent \textbf{Sequence generation} \quad \textcolor{black}{Our task is related to sequence generation~\citep{bahdanau2017actor, yu2017seqgan, paulus2017deep}. \citep{bahdanau2017actor} uses actor-critic for machine translation. Their actor and critic have the same encoder-decoder architecture while the critic outputs state-action value function for each possible actions for policy iteration. Given that the action space is huge for a sequence generation task, the predicted action-value function often have to rely on various tricks to penalising the variance of the outputs of the critic. Without the penalty the values of rare actions can be severely overestimated, introducing bias to the gradient estimates and causing convergence difficulties.} \citep{yu2017seqgan} uses Monte Carlo rollout to sample actions and uses GAN to compute reward. ~\citep{paulus2017deep} applies the same method as~\citep{rennie2016self} to improve the performance of abstractive summarisation.

\section{Methodology}

		\begin{figure*}
	\centering
		\includegraphics[height=5.67cm]{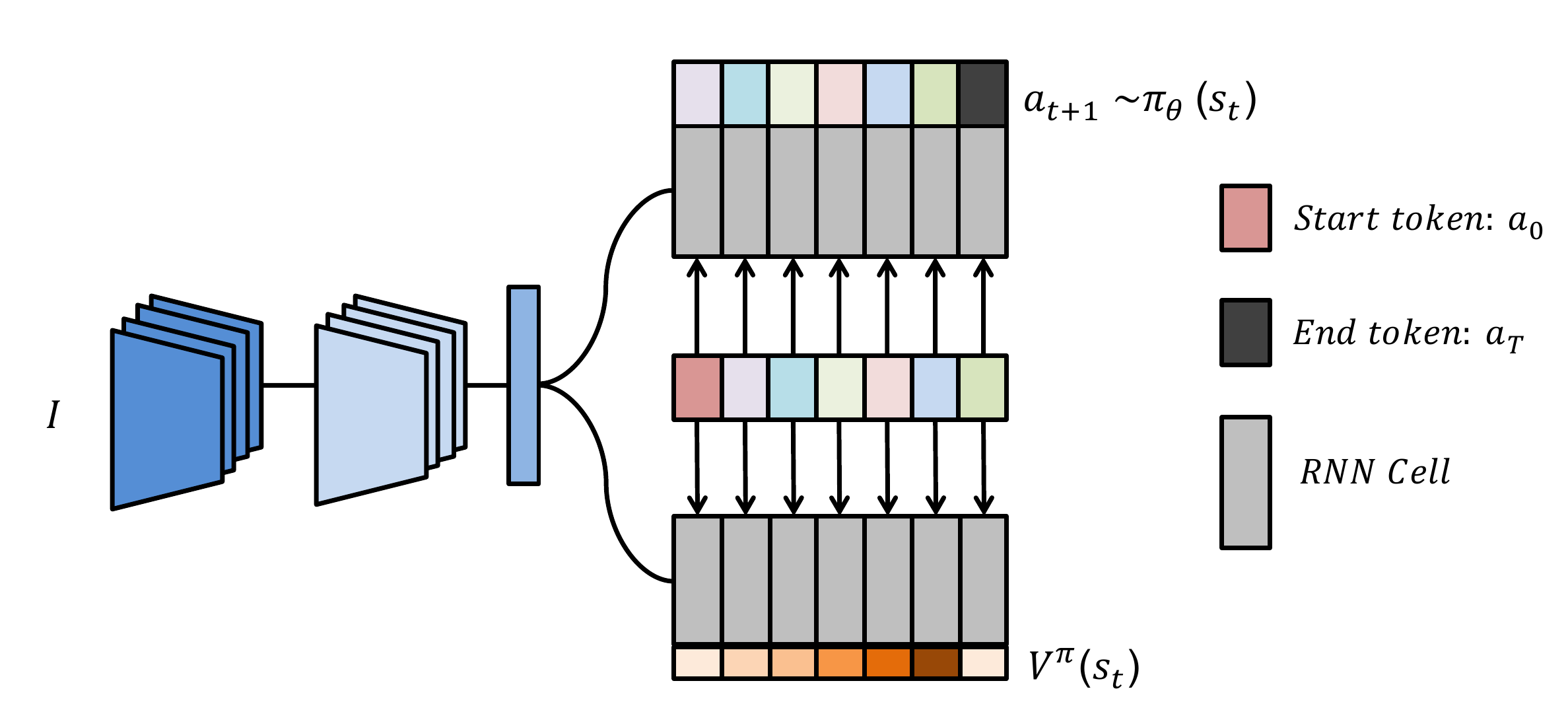}
		\caption{Schematic illustration of our actor-critic based captioning model (with word embedding layer omitted).}
		\label{fig:full_model}
	\end{figure*}

\subsection{Problem formulation}

We consider the problem of learning to generate caption sequence $Y = \{y_{1}, \dots, y_{T}\}, y_{t} \in \mathcal{D}$ given an image $I$, where $\mathcal{D}$ is the dictionary. To simplify the formulas we always use $T$ to denote the length of an output sequence, ignoring the fact that the generated caption sequences may have different lengths. Two sets of input-output pairs $(I, Y)$ are assumed to be available for both training and testing. The trained sequence generative model is evaluated by computing the task-specific score $R(\hat{Y}, Y)$ (e.g., BLEU, CIDEr) on the test set, where $\hat{Y}$ is the predicted caption sequence.

In this work we utilise the conventional encoder-decoder architecture (see Figure \ref{fig:full_model}) for image captioning which consists of a Convolutional Neural Network (CNN) as the encoder and a Recurrent Neural Network (RNN) as the decoder. In order to transform the image captioning problem into a reinforcement learning task, we consider the image caption generation process as a finite Markov decision process (MDP) $\{S, A, P, R, \gamma\}$. In the MDP setting, the state $S$ is composed of the image feature $\mathrm{I}_e$ encoded by the CNN from image $I$ and the tokens/actions $\{a_0, a_1, \dots, a_t\}$ that are generated so far. With the definition of the state, the state transition function $P$ is $s_{t+1} = \{s_{t},~a_{t+1}\}$, where the action $a_{t+1}$ becomes a part of the next state $s_{t+1}$ and the reward $r_{t+1}$ is received. $\gamma \in [0,1]$ is the discount factor. Under the MDP interpretation of the image captioning problem, we can apply standard reinforcement learning algorithms to maximise the cumulated reward. 

\subsection{Model}
We train our model using actor-critic~\citep{Barto1983Neuronlike} reinforcement learning method which contains a policy network (actor) and a value network (critic). In particular, we use Inception-V3~\citep{szegedy2016rethinking} as the CNN subnet and Long Short-Term Memory~\citep{hochreiter1997long} as the RNN subnet. Both the policy network and value network are based on LSTM for sequential action or value  generation.

\noindent \textbf{Policy network} \quad The policy network $\pi$ is parametrised by $\theta$ and at time $t$ it receives a state $s_{t}$ and generates the categorical distribution over $|\mathcal{D}|$ actions (tokens), i.e. $a_{t+1} \sim \pi_{\theta}(s_{t})$. We encode the given image $I$ to $\mathrm{I}_e$ by CNN and treat $\mathrm{I}_e$ and the start token $a_0$ as the initial state $s_0$:

\begin{equation}
	s_0 = \{\mathrm{I}_e, ~a_0\}.
\end{equation}

With state transfer function mentioned above, we have:
\begin{equation}
	s_t = \{\mathrm{I}_e, a_0, a_1, \dots , a_{t}\}.
\end{equation}
We feed state $s_{t}$ into the LSTM and obtain the LSTM hidden state $\mathrm{h}_{t+1}$ ($\mathrm{I}_e$ was set as $\mathrm{h}_{0}$). In order to build a probabilistic model for caption generation with an LSTM, we add a stochastic output layer $f$ (typically with the softmax activation for discrete outputs) that generates outputs $a_{t+1} \in \mathcal{D}$:
\begin{align*}
	\mathrm{h}_{t+1} &= \mathrm{LSTM}(s_t), \\
	a_{t+1} &\sim f(\mathrm{h}_{t+1}).
\end{align*}
Thus, the policy network defines a probability distribution $p(a_{t+1}|s_{t})$ of the action $a_{t+1}$ given current state ${s_t}$. The architecture of the policy network is same as the standard supervised learning. Therefore, by given a target ground truth sequence $\{y_{0},y_{1}, \dots, y_{T}\}$, the supervised learning approach would be to train this network by minimising the cross entropy loss (XE).
\begin{equation}\label{loss:ce}
\mathcal{L}_{\mathrm{XE}}(\theta) = - \sum^{T}_{t=1} \log (\pi_{\theta}(y_{t}~|~y_{0}, \dots, y_{t-1})).
\end{equation}
This corresponds to imitation learning of a perfect teacher in a RL context, and we use the pre-trained model as the initial policy network.

\noindent \textbf{Policy gradient training}\quad 
Policy gradient methods maximise the expected cumulated reward by repeatedly estimating the gradient $g := \nabla_{\theta} \mathbb{E} [\sum^{T}_{t=1}r_{t}]$, where the environment issues the reward $r_{t}$ according to the efficacy of the produced actions, rather than the teacher demonstrating the ideal actions directly as in Eq~\ref{loss:ce}. For policy gradient, it is typically better to train an expression of the form:
\begin{equation}
g = \mathbb{E} [\sum^{T-1}_{t=0}A^{\pi}(s_{t}, ~a_{t+1}) \nabla_{\theta} \log \pi_{\theta} (a_{t+1}~|~s_{t})],     
\end{equation}
where $A^{\pi}(s_{t}, ~a_{t+1})$  is advantage function yields almost the lowest possible variance, though in practice, the advantage function is not known and must be estimated. This statement can be intuitively justified by the following interpretation of the policy gradient: that a step in the policy gradient direction should increase the probability of better-than-average actions and decrease the probability of worse-than-average actions. The advantage function, by it’s definition $A^{\pi}(s, a) = Q^{\pi}(s, a) - V^{\pi}(s)$, measures whether or not the action is better or worse than the policy's default behaviour. So that the gradient term $A^{\pi}(s_{t}, ~a_{t+1}) \nabla_{\theta} \log \pi_{\theta} (a_{t+1}~|~s_{t}) $ points in the direction of increased $\pi_{\theta} (a_{t+1}~|~s_{t})$ if and only if $A^{\pi}(s_{t},~a_{t+1}) > 0$.

\noindent \textbf{Value network} \quad Given the policy $\pi$, sampled actions and reward function, the value represents the expected future return as a function of the observed state $s_{t}$. We use $V$ be an approximate state-value function. 
\begin{equation}
V^{\pi}(s_{t}) = \mathbb{E}[\sum^{T-t-1}_{l=0}\gamma^{l}r_{t+l+1}~|~a_{t+1}, \dots, a_{T} \sim \pi, ~I],
\end{equation}
where discount factor $\gamma$ allows us to reduce variance by down-weighting rewards, at the cost of introducing bias. This parameter corresponds to the discount factor used in discounted formulations of MDPs.

Our value network can be seen as an encoder. We propose to use a separate LSTM parametrised by $\phi$ with shared CNN. The RNN consumes state $s_t = \{\mathrm{I}_e, a_0, a_1, \dots , a_{t}\}$ and produces a single value output to predict the TD target (to be defined later in Sec~\ref{TD}).


\subsection{Advantage function estimation}
\label{TD}
We use temporal-difference (TD) learning for advantage function estimation. Specifically,  we define $Q^{\pi}(s_t, a_{t+1})$ in forward-view TD($\lambda$) setting:
\begin{equation}
	Q^{\pi}(s_t, a_{t+1}) = (1-\lambda) \sum^{\infty}_{n=1}G^n_t,
\end{equation}
where $G^n_t$ is the n-step expected return:
\begin{equation}
	G^n_t = r_{t+1} + \gamma r_{t+2} + ... + \gamma^{n-1}r_{t+n} + \gamma^{n}V^{\pi}(s_{t+n}).
\end{equation}
Therefore we have:
\begin{equation}
	A^{\pi}(s_t,a_{t+1}) = Q^{\pi}(s_t, a_{t+1}) - V^{\pi}(s_t) = (1-\lambda) \sum^{\infty}_{n=1}G^n_t - V^{\pi}(s_t),
\end{equation}
which is the same definition as GAE~\citep{schulman2015high} but in a forward view. Then the gradient of policy network has the form:

\begin{equation}
g = \mathbb{E} [\sum^{T-1}_{t=0}\big((1-\lambda) \sum^{\infty}_{n=1}G^n_t - V^{\pi}(s_t)) \nabla_{\theta} \log \pi_{\theta} (a_{t+1}| s_{t})].    
\end{equation}

\subsection{Value function estimation}
When using a nonlinear function approximator to represent the value function, the simplest approach is to solve a nonlinear regression problem:

\begin{equation}\label{loss:critic}
\min_{\phi}|| Q^{\pi}(s_t,a_{t+1}) - V^{\pi}_{\phi}(s_{t})  ||^{2}
\end{equation}
where $Q^{\pi}(s_t,~a_{t+1}) = (1-\lambda) \sum^{\infty}_{n=1}G^n_t$. 

\subsection{$\lambda$ setting for image captioning}

$\lambda$ setting plays an important role for the whole algorithm. If $\lambda = 0$, the advantage and value function estimations become one-step TD, whereas if $\lambda = 1$, the estimations turn out to be Monte Carlo approach. Since the episode length of image captioning is relatively shorter than popular contemporary RL problems (e.g. Atari and Mujoco games), and we have to sample the whole sequence of captions for rewarding, we set $\lambda = 1$  for our image captioning problem. Under this setting, the estimator for both advantage and value function is unbiased and the limited length of episode restricts the variance of estimation to a limited range.
	Concretely, with $\lambda = 1$, we have:
	\begin{equation}\label{eq:TD}
	Q^{\pi}(s_t,~a_{t+1}) = \sum^{T-t-1}_{l=0} \gamma^{l} r_{t+l+1}
\end{equation}

\subsection{Reward}

For image captioning we can only obtain an evaluation score (e.g.~CIDEr) when the caption generation process is finished. Therefore, we define the reward  as follows:

\begin{equation}
	r_t = \begin{cases}0 & t < T\\ \mathrm{score} &  t = T\end{cases}
\end{equation}

under such reward setting, we have

\begin{equation}
	Q^{\pi}(s_t,~a_{t+1}) = \gamma^{T-t-1} r_T.
\end{equation}

Then the gradient of policy network has a sample form:

\begin{equation}\label{loss:actor}
g = \mathbb{E} [\sum^{T-1}_{t=0}\big(\gamma^{T-t-1} r_T - V(s_t)) \nabla_{\theta} \log \pi_{\theta} (a_{t+1}~|~s_{t})].    
\end{equation}

\section{Experiments}
\label{sec:result}

\subsection{Implementation details}

We use Inception-v3~\citep{szegedy2016rethinking} as the CNN subnet, and an LSTM network is used as the RNN subnet. The number of LSTM cells is 512, equalling to the dimension of the word embedding. The output vocabulary size for sentence generation is 12,000. Note that all these are exactly the same as the NICv2~\citep{vinyals2015show} model ensuring a fair comparison.

For the CNN feature we used,  semantic concept~\citep{liu2016semantic} feature $\mathrm{I} \in \mathbb{R}^{1000}$ is used. These 1,000 semantic concepts are mined from the most frequents words in a set of image captions. A concept classifier is learned to predict Is as classification scores for the concepts.

Algorithm~\ref{algo:training} describes the proposed method in detail. Our preliminary experiments show that training actor-critic from scratch can lead to an early determinisation of the policy and vanishing gradients, because neither the actor nor the critic would provide adequate training signals for one another. The actor would sample completely random tokens that receive very low reward, thus providing a very weak training signal for the critic. A random critic would be similarly useless for training the actor. To overcome these problems, staged pre-trainings are carried out. More specifically, we first pre-train the actor using standard cross entropy loss (XE) (see Eq.~\ref{loss:ce}). After that, we pre-train the critic network by feeding it with sampled actions from the fixed pre-trained actor. The critic network is pre-trained for 2,000 iterations using Adam with a learning rate of 5e-5. For the final stage of joint training of actor-critic, we weight critic loss by 0.5. We use Adam with an initial learning rate of 5e-5 and decrease it to 5e-6 after 1 million iterations, with minibatch size 16. We set discount factor $\gamma = 1$ in all our experiments. The complete training procedure including pre-training is described by Algorithm~\ref{algo:complete}.

\IncMargin{1.5em}
\begin{algorithm}
	\textbf{Require:} Actor $\pi(a_{t+1}| s_{t})$ and critic $V(s_{t})$ with weights $\theta$ and $\phi$ respectively\;
	\For{episode = 1 to max episode}{
		Receive a random example $(I, Y)$
		and sample sequence of actions $\{a_{1}, \dots, a_{T} \} $ according to current policy $\pi_{\theta}$\;
		Compute TD target $Q^{\pi}(s_t,~a_{t+1}) = \gamma^{T-t-1} r_T$ for $V(s_{t})$\;
		Update critic weights $\phi$ by minimising Eq. \ref{loss:critic}\;
		Update actor weights $\theta$ using the gradient in Eq. \ref{loss:actor}\;
	}
	\caption{Actor-Critic Training for Image Captioning}
	\label{algo:training}
\end{algorithm}
\DecMargin{1.5em}


\IncMargin{1.5em}
\begin{algorithm}
	Initialise actor  $\pi(a_{t+1}| s_{t})$  and critic $V(s_{t})$ with random weights $\theta$ and $\phi$ respectively\;
	Pre-train the actor to predict ground truth $y_{t}$ given $\{y_{1}, \dots, y_{t-1}\}$ by minimise Eq. \ref{loss:ce}\;
	Pre-train the critic to estimate $V(s_{t})$ by running Algorithm \ref{algo:training} with fixed actor\;
	Run Algorithm \ref{algo:training}
	\caption{Complete Actor-Critic Algorithm for Image Captioning}
	\label{algo:complete}
\end{algorithm}
\DecMargin{1.5em}

\subsection{Datasets and setting}
We evaluate the  proposed method on the most widely used MSCOCO \cite{lin2014microsoft} dataset. The dataset contains 82,783 training images and 40,504 validation images. Each image is manually annotated with about 5 captions. The comparison against the state-of-the-art is conducted using the actual MS COCO test set comprising 40,775 images. Note that the annotation of the test set is not publicly available, so the results are obtained from the COCO evaluation server. We also follow the setting of \cite{vinyals2015show, vinyals2016show} by using a held-out set of 4,051 images from the COCO validation set as the development set. The widely used BLEU, CIDEr, METEOR, and ROUGE scores are employed to measure the quality of generated captions.

\subsection{Experimental results}
\noindent \textbf{Competitors} \quad Several state-of-the-art models are selected for comparison: MSRCap: The Microsoft Captivator~\citep{devlin2015language} combines the bottom-up based word generation model~\citep{fang2015captions} with a gated recurrent neural network~\citep{cho2014learning} (GRNN) for image captioning. mRNN: The multimodal recurrent neural network~\citep{mao2014deep} uses a multimodal layer to combine the CNN and RNN. NICv2: The NICv2~\citep{vinyals2016show} is an improved version of the Neural Image Caption generator~\citep{vinyals2015show}. It uses a better image encoder, i.e., Inception-v3. In addition, scheduled sampling~\citep{bengio2015scheduled} and an ensemble of 15 models are used; both improved the accuracy of captioning. V2L: The V2L model~\citep{wu2016value} uses a CNN based attribute detector to firstly generate 256 attributes, and then feed as initial input to an LSTM model to generate captions. ATT: The semantic attention model~\citep{you2016image} uses both image features and visual attributes, and introduces an attention mechanism to reweight the attribute context to improve captioning accuracy. Semantic~\citep{liu2016semantic} is our base model which uses a semantically regularised embedding layer as the interface between the CNN and RNN. 

In addition to the traditional supervised learning method, we compare our method with three reinforcement learning based model. PG~\citep{liu2016optimization} and MIXER~\citep{ranzato2015sequence}  use policy gradient method with an additional FC layer on top of the RNN as state value network to reduce the high variance. Different from~\citep{ranzato2015sequence},~\citep{liu2016optimization} uses the same method as   \citep{yu2017seqgan} with $k$-times Monte Carlo rollout to estimate the state value target. Self-critical~\citep{rennie2016self} uses the basic REINFORCE algorithm with a reward obtained by the current model under the inference algorithm as the baseline. This simple method achieved very high performance which ranked 2nd currently on coco captioning challenge. However, it use a multiple model ensemble. In contrast, only a single model is used for our method. 

\begin{figure*}
	\centering
		\includegraphics[height=5.67cm]{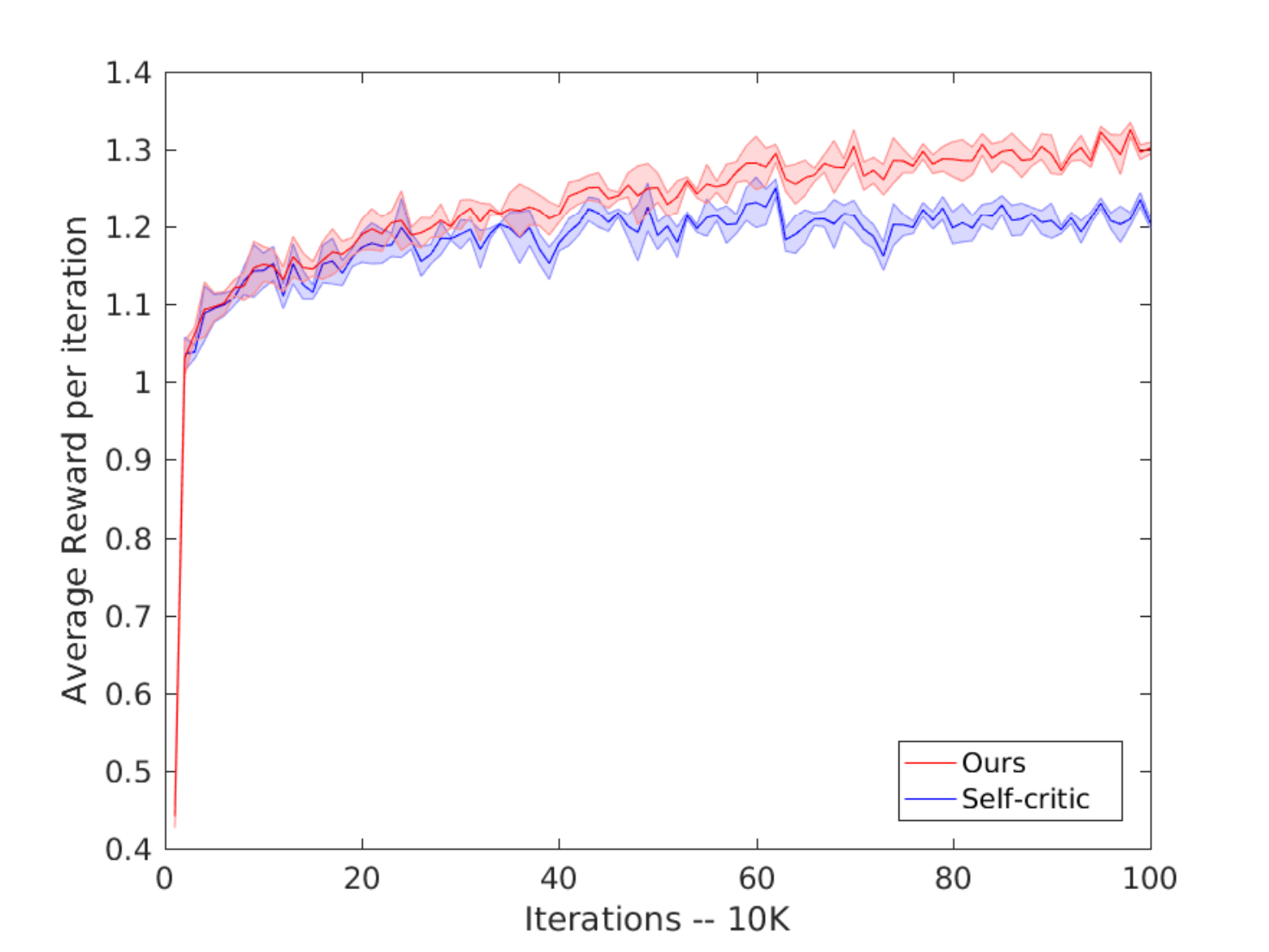}
		\caption{Average training reward curve for ours and~\citep{rennie2016self}. We recorded the reward for 1 million iteration and plot every 10k iteration.}
		\label{fig:reward}
	\end{figure*}

\noindent \textbf{Results} \quad The results on development set are summarised in Table \ref{tab:dev}. We report a significant improvement from 1.007 to 1.162 on CIDEr over the log-likelihood baseline when single model greedy search is used for decoding. We can also see that our method is better than attention~\cite{xushow} and memory cell~\citep{graves2016hybrid} which are added on top of the LSTM cell. For fair comparison with the current state-of-the-art method~\citep{rennie2016self}, we implement it with same semantic CNN input~\citep{liu2016semantic} on development set with single model for evaluation. The training reward curves of our method and ~\citep{rennie2016self} are shown in Figure~\ref{fig:reward}. Our method is clearly superior to that of \citep{rennie2016self} due to the per-token advantage and value computation strategy adopted in our actor-critic based reinforcement learning framework.

\setlength{\tabcolsep}{5pt}
\begin{table*}[htbp]
\begin{center}
\footnotesize
\begin{tabular}{@{} l|c|c c c c@{}}
\toprule
{\textbf{Metric}} &\textbf{CIDEr-D} &\textbf{BLEU-4} &\textbf{METEOR}&\textbf{ROUGE-L}\\ 

\midrule 
NIC~\citep{vinyals2015show} & 0.855 & 0.277 &0.237 &-\\ 
NICv2~\citep{vinyals2016show} & 0.998 & 0.321 &0.257 &-\\ 
Semantic~\citep{liu2016semantic} &1.007 &0.302 &0.256 &0.539\\ 
Semantic~\citep{liu2016semantic}+Attention~\cite{xushow} &1.042 &0.311 &0.263 &0.543\\
Semantic~\citep{liu2016semantic}+Attention~\cite{xushow}+Memory~\citep{graves2016hybrid} &1.057&0.318  &0.266 &0.547\\ 
Semantic~\citep{liu2016semantic}+Self-critical~\citep{rennie2016self} &1.140&0.323 &0.266 &0.554\\
\midrule
Ours  &\bf 1.162 &\bf 0.344 &\bf 0.267 &\bf 0.558\\ 
\bottomrule
\end{tabular}
\end{center}
\caption{Single model greedy search  scores on the MSCOCO development set}
\label{tab:dev}
\end{table*}

\begin{figure}[!htb]
\begin{center}
\begin{tabular}{@{}l@{}l@{}l}
\includegraphics[width=0.30\linewidth]{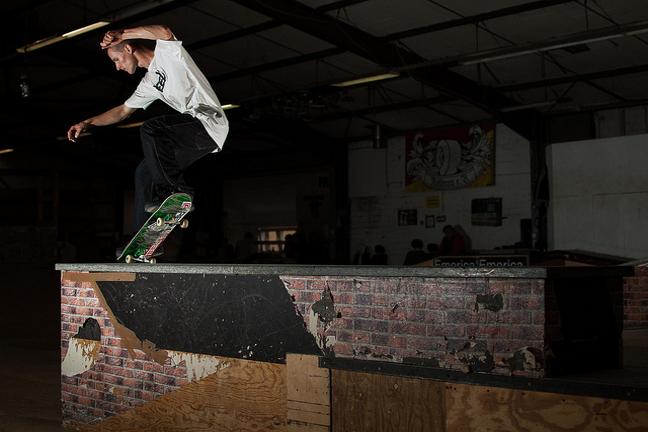}
& \includegraphics[width=0.30\linewidth]{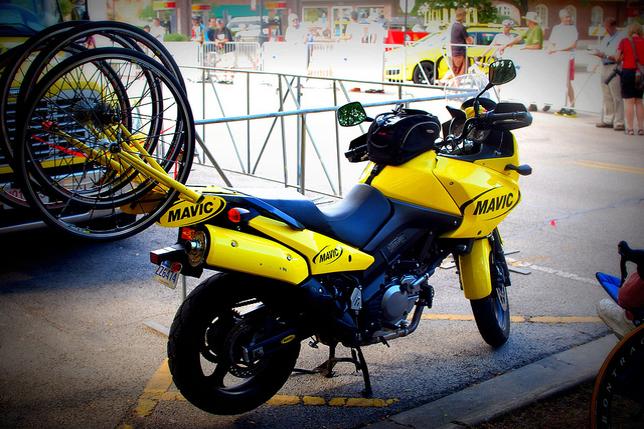}
& \includegraphics[width=0.30\linewidth]{./321557.jpg}\\
\tiny Human	: A man doing a trick on this skateboard. \quad \quad \quad \quad
 & \tiny Human	: A motorcycle carrying many wheels is parked.  
 & \tiny Human	: Large black motorcycle sitting next to a white building. \\
\tiny XE	: a man riding a skateboard on a cement ledge. 
 & \tiny XE	: a motorcycle parked next to a {\color{red} yellow wall}. 
 & \tiny XE	: a motorcycle parked next to a building. \\
\tiny Ours	: a man doing a {\color{green} trick} on a skateboard. 
 & \tiny Ours	: a {\color{green} yellow motorcycle} parked in front of a street. 
 & \tiny Ours	: a {\color{green} black motorcycle} parked next to a building. \\
\includegraphics[width=0.30\linewidth]{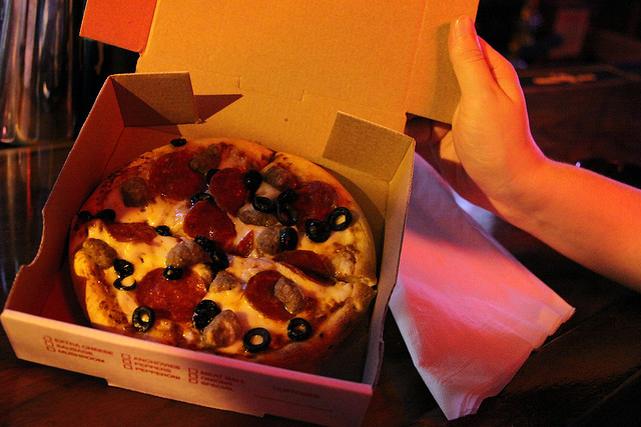}
& \includegraphics[width=0.30\linewidth]{./150314.jpg}
& \includegraphics[width=0.30\linewidth]{./439589.jpg}\\
\tiny Human	: A small personal pizza sits in a pizza box. 
 & \tiny Human	: A group of skiers in the mountains reach a sign. 
 & \tiny Human	: Old and new trains navigating a rail yard. \\
\tiny XE	: a pizza is {\color{red} sitting} on a box with a box of pizza. 
 & \tiny XE	: a group of people {\color{red} standing} on top of slope. 
 & \tiny XE	: a train is on the tracks in a city. \\
\tiny Ours	: a person {\color{green} holding} a box of pizza. 
 & \tiny Ours	: a group of people {\color{green} skiing} on a snow covered slope. 
 & \tiny Ours	: a {\color{green} black and white photo} of trains on a train yard. \\
\end{tabular}
\end{center}
\caption{Qualitative results of image captioning on the MS COCO dataset. } \label{fig:capgood}
\end{figure}

We also submitted our single model results to the official evaluation server to compare with the eight baselines mentioned above. The evaluation is done with both 5 and 40 reference captions (C5 and C40). Our model is ranked the 3rd on the  MSCOCO image captioning challenge leaderboard when submitted. Table \ref{tab:learderboard} shows that, compare to the supervised learning based methods, our approach significantly outperforms all of them in all metrics despite using only a single model rather than model ensemble. Comparing to the other two reinforcement learning based methods~\citep{liu2016optimization, ranzato2015sequence}, our method still achieved better performance except ROUGE-L c40. Figure \ref{fig:capgood} shows some qualitative examples of our models captioning compared against ground truth (Human) and method using the same encoder-decoder architecture, but with standard cross-entropy (XE) training.

\setlength{\tabcolsep}{2pt}
\begin{table*}[htbp]
\begin{center}
\scriptsize
\begin{tabular}{p{1.80cm} |p{0.7cm}<{\centering} p{0.7cm}<{\centering} | p{0.7cm}<{\centering} p{0.7cm}<{\centering} | p{0.7cm}<{\centering} p{0.7cm}<{\centering} | p{0.7cm}<{\centering} p{0.7cm}<{\centering} | p{0.7cm}<{\centering} p{0.7cm}<{\centering} | p{0.7cm}<{\centering} p{0.7cm}<{\centering} | p{0.7cm}<{\centering} p{0.7cm}<{\centering}}
\toprule
\multirow{2}{*}{\bf Metric} 	& \multicolumn{2}{c}{\bf B-1}	\vline &  \multicolumn{2}{c}{\bf B-2} \vline 	& \multicolumn{2}{c}{\bf B-3} \vline	& \multicolumn{2}{c}{\bf B-4} \vline	& \multicolumn{2}{c}{\bf METEOR} \vline	& \multicolumn{2}{c}{\bf ROUGE-L} \vline & \multicolumn{2}{c}{\bf CIDEr-D}   \\ 
 	& c5	& c40	& c5	& c40	& c5	& c40	& c5	& c40	& c5	& c40	& c5	& c40	& c5	& c40 \\ 
\toprule
MSRCap~\citep{devlin2015language}	& 0.715	& 0.907	& 0.543	& 0.819	& 0.407	& 0.710	& 0.308& 0.601	& 0.248	& 0.339	& 0.526& 0.680	& 0.931	& 0.937 \\ 
mRNN~\citep{mao2014deep}	& 0.716	& 0.890	& 0.545	& 0.798	& 0.404	& 0.687	& 0.299	& 0.575	& 0.242	& 0.325	& 0.521	& 0.666	& 0.917	& 0.935 \\ 
V2L~\citep{wu2016value} & 0.725	& 0.892	& 0.556	& 0.803	& 0.414	& 0.694& 0.306	& 0.582	& 0.246	& 0.329	& 0.528	& 0.672	& 0.911	& 0.924 \\ 
NICv2~\citep{vinyals2016show}	& 0.713	& 0.895	& 0.542	& 0.802& 0.407	& 0.694	& 0.309& 0.587	& 0.254	& 0.346	& 0.530	& 0.682	& 0.943	& 0.946 \\ 
ATT~\citep{you2016image}	& 0.731	& 0.900	& 0.565	& 0.815	& 0.424	& 0.709& 0.316	& 0.599	& 0.250	& 0.335	& 0.535	& 0.682	& 0.943	& 0.958 \\ 
Semantic~\citep{liu2016semantic}	& 0.743	& 0.917	& 0.578& 0.840	& 0.434	& 0.735& 0.323	& 0.621	& 0.255	& 0.343	& 0.540	& 0.691& 0.986	& 1.002 \\ 
 
PG~\citep{liu2016optimization}	& 0.754	& 0.918	& 0.591& 0.841	& 0.445	& 0.738& 0.332	& 0.624	& 0.257	& 0.340	& 0.550	& \textbf{0.695}& 1.013	& 1.032 \\ 
MIXER~\citep{ranzato2015sequence}	& 0.747	& -	& 0.579& -	& 0.431	& -& 0.317	&- & 0.258	& -	& 0.545	& -& 0.991	& - \\ 
\midrule
Ours	& \textbf{0.778}	& \textbf{0.929}	& \textbf{0.612}	& \textbf{0.855}	& \textbf{0.459}	& \textbf{0.745}	& \textbf{0.337}	& \textbf{0.625}	& \textbf{0.264}	& \textbf{0.344}	&\textbf{0.554}	&{0.691}	& \textbf{1.102}& \textbf{1.121} \\ 
\bottomrule
\end{tabular}
\end{center}
\vspace{-1mm}
\caption{Results from the official MS-COCO image captioning challenge leaderboard (\url{http://cocodataset.org/\#captions-leaderboard})}
\label{tab:learderboard}
\end{table*}

\noindent \textbf{Computational Cost}  We compare the training time of our method with several alternatives. All algorithms are implemented in \emph{Tensorflow} and run on an NVIDIA P100 card, with a mini-batch size of 16. Table~\ref{tab:time} shows that for training, our method is the most efficient one. This is mainly due to the fact that our model does not have attention cell. Furthermore, for the Self-critical~\citep{rennie2016self} model, it needs to \textcolor{black}{sample twice} (random sampling + greedy decoding) for each iteration which is expensive.

\setlength{\tabcolsep}{10pt}
\begin{table}[htbp]
\begin{center}
\footnotesize
\begin{tabular}{@{} l|c @{}}
\toprule
\textbf{Model} &\textbf{Time}  \\ 
\midrule 
Semantic~\citep{liu2016semantic}+Attention~\cite{xushow}&  0.10\\ 
Semantic~\citep{liu2016semantic}+Self-critical~\citep{rennie2016self}& 0.13 \\ 
Semantic~\citep{liu2016semantic}+Self-critical~\citep{rennie2016self}+Attention~\cite{xushow}& 0.18\\ 
\midrule 
Ours  &\bf 0.07    \\ 
\bottomrule
\end{tabular}
\end{center}
\caption{Training time for one minibatch on COCO dataset (in seconds)}
\label{tab:time}
\end{table}

\section{Conclusion}
\label{sec:conclusion}
We have investigated the problem of  automated image captioning by employing reinforcement learning to optimise the relevant non-differentiable language metrics such as CIDEr. A novel actor-critic based learning strategy is formulated which has the advantage over existing reinforcement learning based captioning models in that a per-token advantage and value computation is enabled leading to better model training. State-of-the-art performance is achieved using our computational efficient model on the MSCOCO benchmark.

\bibliographystyle{plain}
\bibliography{caption}

\begin{thebibliography}{10}

\bibitem{bahdanau2017actor}
Dzmitry Bahdanau, Philemon Brakel, Kelvin Xu, Anirudh Goyal, Ryan Lowe, Joelle
  Pineau, Aaron Courville, and Yoshua Bengio.
\newblock An actor-critic algorithm for sequence prediction.
\newblock In {\em ICLR}, 2017.

\bibitem{Barto1983Neuronlike}
A.~G. Barto, R.~S. Sutton, and C.~W. Anderson.
\newblock Neuronlike adaptive elements that can solve difficult learning
  control problems.
\newblock {\em IEEE Transactions on Systems, Man, and Cybernetics}, 1983.

\bibitem{bengio2015scheduled}
Samy Bengio, Oriol Vinyals, Navdeep Jaitly, and Noam Shazeer.
\newblock Scheduled sampling for sequence prediction with recurrent neural
  networks.
\newblock In {\em NIPS}, 2015.

\bibitem{cho2014learning}
Kyunghyun Cho, Bart Van~Merri{\"e}nboer, Caglar Gulcehre, Dzmitry Bahdanau,
  Fethi Bougares, Holger Schwenk, and Yoshua Bengio.
\newblock Learning phrase representations using rnn encoder-decoder for
  statistical machine translation.
\newblock {\em arXiv preprint arXiv:1406.1078}, 2014.

\bibitem{devlin2015language}
Jacob Devlin, Hao Cheng, Hao Fang, Saurabh Gupta, Li~Deng, Xiaodong He,
  Geoffrey Zweig, and Margaret Mitchell.
\newblock Language models for image captioning: The quirks and what works.
\newblock In {\em ACL}, 2015.

\bibitem{fang2015captions}
Hao Fang, Saurabh Gupta, Forrest Iandola, Rupesh~K Srivastava, Li~Deng, Piotr
  Doll{\'a}r, Jianfeng Gao, Xiaodong He, Margaret Mitchell, John~C Platt,
  et~al.
\newblock From captions to visual concepts and back.
\newblock In {\em CVPR}, 2015.

\bibitem{graves2016hybrid}
Alex Graves, Greg Wayne, Malcolm Reynolds, Tim Harley, Ivo Danihelka, Agnieszka
  Grabska-Barwi{\'n}ska, Sergio~G{\'o}mez Colmenarejo, Edward Grefenstette,
  Tiago Ramalho, John Agapiou, et~al.
\newblock Hybrid computing using a neural network with dynamic external memory.
\newblock {\em Nature}, 2016.

\bibitem{hochreiter1997long}
Sepp Hochreiter and J{\"u}rgen Schmidhuber.
\newblock Long short-term memory.
\newblock {\em Neural computation}, 1997.

\bibitem{lin2014microsoft}
Tsung-Yi Lin, Michael Maire, Serge Belongie, James Hays, Pietro Perona, Deva
  Ramanan, Piotr Doll{\'a}r, and C~Lawrence Zitnick.
\newblock Microsoft coco: Common objects in context.
\newblock In {\em ECCV}, 2014.

\bibitem{liu2016semantic}
Feng Liu, Tao Xiang, Timothy~M Hospedales, Wankou Yang, and Changyin Sun.
\newblock Semantic regularisation for recurrent image annotation.
\newblock In {\em CVPR}, 2017.

\bibitem{liu2016optimization}
Siqi Liu, Zhenhai Zhu, Ning Ye, Sergio Guadarrama, and Kevin Murphy.
\newblock Improved image captioning via policy gradient optimization of spider.
\newblock In {\em ICCV}, 2017.

\bibitem{mao2014deep}
Junhua Mao, Wei Xu, Yi~Yang, Jiang Wang, Zhiheng Huang, and Alan Yuille.
\newblock Deep captioning with multimodal recurrent neural networks (m-rnn).
\newblock In {\em ICLR}, 2015.

\bibitem{mnih2016asynchronous}
Volodymyr Mnih, Adria~Puigdomenech Badia, Mehdi Mirza, Alex Graves, Timothy~P
  Lillicrap, Tim Harley, David Silver, and Koray Kavukcuoglu.
\newblock Asynchronous methods for deep reinforcement learning.
\newblock In {\em ICML}, 2016.

\bibitem{mnih2013playing}
Volodymyr Mnih, Koray Kavukcuoglu, David Silver, Alex Graves, Ioannis
  Antonoglou, Daan Wierstra, and Martin Riedmiller.
\newblock Playing atari with deep reinforcement learning.
\newblock In {\em NIPS}, 2013.

\bibitem{paulus2017deep}
Romain Paulus, Caiming Xiong, and Richard Socher.
\newblock A deep reinforced model for abstractive summarization.
\newblock {\em arXiv preprint arXiv:1705.04304}, 2017.

\bibitem{ranzato2015sequence}
Marc'Aurelio Ranzato, Sumit Chopra, Michael Auli, and Wojciech Zaremba.
\newblock Sequence level training with recurrent neural networks.
\newblock In {\em ICLR}, 2016.

\bibitem{rennie2016self}
Steven~J Rennie, Etienne Marcheret, Youssef Mroueh, Jarret Ross, and Vaibhava
  Goel.
\newblock Self-critical sequence training for image captioning.
\newblock In {\em CVPR}, 2017.

\bibitem{schulman2015high}
John Schulman, Philipp Moritz, Sergey Levine, Michael Jordan, and Pieter
  Abbeel.
\newblock High-dimensional continuous control using generalized advantage
  estimation.
\newblock In {\em ICLR}, 2016.

\bibitem{silver2016mastering}
David Silver, Aja Huang, Chris~J Maddison, Arthur Guez, Laurent Sifre, George
  Van Den~Driessche, Julian Schrittwieser, Ioannis Antonoglou, Veda
  Panneershelvam, Marc Lanctot, et~al.
\newblock Mastering the game of go with deep neural networks and tree search.
\newblock {\em Nature}, 2016.

\bibitem{szegedy2016rethinking}
Christian Szegedy, Vincent Vanhoucke, Sergey Ioffe, Jon Shlens, and Zbigniew
  Wojna.
\newblock Rethinking the inception architecture for computer vision.
\newblock In {\em CVPR}, 2016.

\bibitem{vedantam2015cider}
Ramakrishna Vedantam, C~Lawrence~Zitnick, and Devi Parikh.
\newblock Cider: Consensus-based image description evaluation.
\newblock In {\em CVPR}, 2015.

\bibitem{vinyals2015show}
Oriol Vinyals, Alexander Toshev, Samy Bengio, and Dumitru Erhan.
\newblock Show and tell: A neural image caption generator.
\newblock In {\em CVPR}, 2015.

\bibitem{vinyals2016show}
Oriol Vinyals, Alexander Toshev, Samy Bengio, and Dumitru Erhan.
\newblock Show and tell: Lessons learned from the 2015 mscoco image captioning
  challenge.
\newblock {\em PAMI}, 2016.

\bibitem{williams1992reinforce}
Ronald~J Williams.
\newblock Simple statistical gradient-following algorithms for connectionist
  reinforcement learning.
\newblock {\em Machine Learning}, 1992.

\bibitem{wu2016value}
Qi~Wu, Chunhua Shen, Lingqiao Liu, Anthony Dick, and Anton van~den Hengel.
\newblock What value do explicit high level concepts have in vision to language
  problems?
\newblock In {\em CVPR}, 2016.

\bibitem{xushow}
Kelvin Xu, Jimmy~Lei Ba, Ryan Kiros, Kyunghyun Cho, Aaron Courville, Ruslan
  Salakhutdinov, Richard~S Zemel, and Yoshua Bengio.
\newblock Show, attend and tell: Neural image caption generation with visual
  attention.
\newblock In {\em ICML}, 2015.

\bibitem{you2016image}
Quanzeng You, Hailin Jin, Zhaowen Wang, Chen Fang, and Jiebo Luo.
\newblock Image captioning with semantic attention.
\newblock In {\em CVPR}, 2016.

\bibitem{yu2017seqgan}
Lantao Yu, Weinan Zhang, Jun Wang, and Yong Yu.
\newblock Seqgan: sequence generative adversarial nets with policy gradient.
\newblock In {\em AAAI}, 2017.

\end{thebibliography}

\end{document}